 \documentclass[pmlr,twocolumn,10pt]{jmlr} 

\usepackage{booktabs}
\usepackage{siunitx}

\newcommand{\N}{\mathbb{N}}
\newcommand{\R}{\mathbb{R}}
\newcommand{\C}{\mathbb{C}}

\newcommand{\dmin}{d_\text{min}}
\newcommand{\dmax}{d_\text{max}}

\renewcommand{\Re}{\mathfrak{Re}}
\renewcommand{\Im}{\mathfrak{Im}}

\theorembodyfont{\upshape}
\theoremheaderfont{\scshape}
\theorempostheader{:}
\theoremsep{\newline}

 \title[Sleep Staging from Airflow Signals Using Fourier Approximations of
 Persistence Curves]{Sleep Staging from Airflow Signals Using Fourier
 Approximations of Persistence Curves}

\author{%
  \Name{Shashank Manjunath} \Email{manjunath.sh@northeastern.edu}\\
  \addr Khoury College of Computer Sciences, Northeastern University, USA
  \AND
  \Name{Hau-Tieng Wu} \Email{hw3635@nyu.edu}\\
  \addr Courant Institute of Mathematical Sciences, New York University, USA
  \AND
  \Name{Aarti Sathyanarayana} \Email{a.sathyanarayana@northeastern.edu}\\
  \addr Khoury College of Computer Sciences and Bouv\'e College of Health Sciences,
  Northeastern University, USA
}

\begin{document}

\maketitle

\begin{abstract}
  Sleep staging is a challenging task, typically manually performed by sleep
  technologists based on electroencephalogram and other biosignals of patients
  taken during overnight sleep studies. Recent work aims to leverage automated
  algorithms to perform sleep staging not based on electroencephalogram signals,
  but rather based on the airflow signals of subjects. Prior work uses ideas from
  topological data analysis (TDA), specifically Hermite function expansions of
  persistence curves (HEPC) to featurize airflow signals. However, finite order
  HEPC captures only partial information. In this work, we propose Fourier
  approximations of persistence curves (FAPC), and use this technique to perform
  sleep staging based on airflow signals. We analyze performance using an XGBoost
  model on 1155 pediatric sleep studies taken from the Nationwide Children's
  Hospital Sleep DataBank (NCHSDB), and find that FAPC methods provide
  complimentary information to HEPC methods alone, leading to a 4.9\% increase
  in performance over baseline methods.
\end{abstract}


\paragraph*{Data and Code Availability}
The data used in this article is available through the National Sleep
Research Resource at \url{https://sleepdata.org/datasets/nchsdb}. The code used
for this article can be found at \url{https://github.com/shashankmanjunath/tda_sleep_staging}.

\paragraph*{Institutional Review Board (IRB)}

All datasets used in this study are publicly available. IRB approval is not
required for this study.

\section{Introduction} \label{sec:intro}

Sleep disorders are a family of common diseases. Approximately 61 \% of the
older adult population experiencing poor sleep, and 3.7 \% of the pediatric
population has an actual diagnosis for a sleep
disorder~\citep{karnaSleepDisorder2024,meltzerPrevalenceDiagnosedSleep2010}. In
order to effectively diagnose sleep disorders, patients typically have to
undergo an overnight sleep study, or polysomnogram
(PSG)~\citep{gerstenslagerSleepStudy2024}. Polysomnograms involve recording a
multitude of sensors, including electroencephalogram, airflow and blood
oxygenation, among others. These PSG exams can be conducted in a clinical
setting at a hospital with sleep technicians watching as the patient sleeps, or
can be self-administered at home while the patient sleeps in their own bed.
At-home PSG studies are convenient and alleviate the need for the patient to
schedule and attend a sleep study in the hospital; however, in some situations
EEG signals may not be available~\citep{campbellEEGRecordingAnalysis2009}. We
thus require methods that can perform sleep staging from non-EEG channels. This
would eliminate the need for EEG sensors in many sleep studies, and especially
in at-home sleep studies.
In this paper, we focus on methods from topological data analysis (TDA) which
have the ability to effectively featurize nonstationary physiological
time-series signals~\citep{zhengTopologicalDataAnalysis2023}. TDA methods have
been shown to effectively analyze complex EEG signals, heart rate variability,
as well as other dynamic and nonstationary
signals~\citep{manjunathTopologicalDataAnalysis2023b,xuTopologicalDataAnalysis2021a,chungTopologicalDataAnalysis2024,chenQualityAwareSleep2023}.
We specifically focus on Persistent Homology (PH), a subset of TDA which has
been applied to study nonstationary time series.

While TDA provides a promising avenue for the analysis of complex time-series
signals, crafting proper features for machine learning purposes remains
challenging. After initial processing, the standard topological descriptor
of data created by PH methods is called the \emph{persistence diagram}
(PD)~\citep{edelsbrunnerComputationalTopologyIntroduction2022}. This
mathematical object is poorly suited to machine learning applications due to its
intricate metric structure. In this work, we focus on \emph{persistence curves}
(PCs),
an efficient approach to PD featurization which leads to an efficient functional
summarization of PDs in a Hilbert
space~\citep{chungPersistenceCurvesCanonical2021}. Prior work proposed using
Hermite function expansions to summarize PCs, called Hermite function Expansions
of Persistence Curves (HEPC)~\citep{chungStableTopologicalFeature2022}. While
HEPC performs well in several applications, in general finite Hermite functions
may miss high frequency features of PCs during approximation. We thus propose
Fourier series based approximations of PCs as an individual feature or as a
supplement to HEPC features in this work.

Our main contributions are:

\begin{enumerate}
  \item We propose Fourier Approximations of Persistence Curves (FAPC), a novel
    framework for approximating PCs which provides complimentary information to
    HEPC and can enhance machine learning model performance
  \item We evaluate FAPC methods for 3-class automated sleep staging
    (Wake/NREM/REM) based on respiratory signals collected during pediatric PSG
    studies in order to explore the potential of this method to help eliminate
    the need for EEG data recording in sleep studies
  \item We test our features with comparison to baseline on 1155 of pediatric
    sleep studies from 1155 unique subjects. Prior work focused on airflow-based
    sleep staging has primarily focused on
    adult sleep studies, so our analysis of respiratory-based sleep staging on
    pediatric sleep studies is novel~\citep{chungTopologicalDataAnalysis2024}.
\end{enumerate}

\section{Related Work}\label{sec:related_work}

\paragraph{Automated Sleep Staging Algorithms}\label{subsec:sleep_staging}

There has been significant recent work in developing automated sleep staging
algorithms, especially algorithms which leverage machine learning techniques for
prediction of sleep stage. Many of these algorithms use EEG signals and
supervised machine learning for sleep stage
classification~\citep{zhangReviewAutomatedSleep}.
Existing work was able to achieve an average accuracy of 91.4\% on a 5-class
sleep stage classification problem using 10 human subjects and an SVM model with
features extracted from EEG
data~\citep{tianHierarchicalClassificationMethod2017}.

\paragraph{Airflow-Based Sleep Staging}\label{subsec:airflow_sleep_staging}

Existing work on an adult cohort of 74 patients was able to achieve a
classification accuracy of 84.8\% using heart rate, respiratory flow, and
respiratory effort signals using a deep learning
model~\citep{bakkerEstimatingSleepStages2021}. Other work uses TDA methods
on airflow signals alone, achieving an accuracy of 69.1 \% on a cohort of 80
adult subjects using an XGBoost model~\citep{chungTopologicalDataAnalysis2024}.
Other airflow-based sleep staging methods on airflow signals alone using non-TDA
methods achieves an accuracy of 77.85 \% on PSG studies from 29 subjects using a
bagging classifier combined with heuristic rules for Wake/REM/NREM
classification~\citep{tataraidzeSleepStageClassification2015}.

\paragraph{Persistence Diagram Featurization Methods}

There has been significant work on featurization of
PDs~\citep{barnesComparativeStudyMachine2021a}. PDs cannot be embedded in a
Hilbert space, which is required by many machine learning
methods~\citep{bubenikStatisticalTopologicalData}. Some existing methods for
representing PDs in Hilbert spaces include persistence landscapes, persistence
images, template systems, and adaptive template
systems~\citep{bubenikStatisticalTopologicalData,adamsPersistenceImagesStable2017,pereaApproximatingContinuousFunctions2023,polancoAdaptiveTemplateSystems2019a}.
Further work includes PCs, which generalize persistence landscapes and allow for
novel insights into datasets which cannot be gleaned directly from persistence
landscapes~\citep{chungPersistenceCurvesCanonical2021}. However, since
PCs are functions, leveraging them in machine learning
algorithms requires computing a large number of mesh points. Recent work has
focused on summarizing PCs using orthogonal polynomials such as Hermite
polynomials~\citep{chungStableTopologicalFeature2022}. This method leverages PCs
while allowing an efficient summarization of the curve with coefficients of an
orthogonal polynomial expansion.

\section{Methods}\label{sec:methods}

\subsection{Dataset and Preprocessing}\label{sec:dataset}

We use Nationwide Childrens Sleep DataBank (NCHSDB) dataset for this
work~\citep{leeLargeCollectionRealworld2022c,zhangNationalSleepResearch2018c}.
This dataset contains 3984 PSG studies from 3673 pediatric subjects. The PSG
studies contain sleep stage labels by sleep technicians for every 30 seconds of
sleep, as well as labels for events during sleep such as oxygen desaturation and
apnea events. For this work on sleep staging using respiratory sensors, we
leverage airway flow as measured by nasal cannula as our respiratory
measure, referred to as ``airflow'' in this paper. We require that subjects used
in our train and test data are less than 18 years old and at least 2 years old
at the time of the study, and that subjects have an apnea-hypopnea index (AHI)
of less than 1, which is considered healthy in pediatric
subjects~\citep{gouthroPediatricObstructiveSleep2024}. We are left with 1155 PSG
studies from 1155 unique subjects. We lastly remove any airflow signal epochs
which are co-located with central, mixed, or obstructive apnea events, hypopnea
events, or oxygen desaturation events. Our airflow signal is sampled at 256 Hz
with 30 second epochs. Detailed sample counts and demographic information about
our selected group of subjects is in Appendix~\ref{apd:demographics}.

We evaluate signal quality by calculating a Signal Quality Index (SQI) on each
epoch of data, following the original formulation by
\citep{birrenkottRespiratoryQualityIndex} and implementation of
\citep{chungTopologicalDataAnalysis2024}.
%
%
If $\text{SQI} \geq 0.25$, we consider the signal to have sufficiently high
quality for downstream use.

\paragraph{Instantaneous Respiratory Rate}


We use the NeuroKit package on our airflow data to perform linear detrending and
apply a 2 Hz low-pass IIR Butterworth
filter~\citep{makowskiNeuroKit2PythonToolbox2021}. We furthermore use NeuroKit
to determine the onset of each breathing cycle. The instantaneous respiratory
rate (IRR) is defined by the monotone cubic spline of the nonuniform $N$
detected breathing cycles. The resulting function is sampled at 4 Hz. The
function represents cycles of breathing per minute. High fluctuation of this
signal indicates high breathing variability.

\paragraph{Classical Features for Analysis}

We construct baseline non-TDA features for comparison with our TDA
features~\citep{tataraidzeSleepStageClassification2015}. These features are
calculated on a the epoch of interest and the previous 5 epochs of data (6
epochs total), consisting of 180 seconds of a time-series signal from the nasal
cannula sensor. We construct 11 features enumerated in
Table~\ref{tab:classic_features}.

\begin{table}[htbp]
  \vspace{-2em}
  \centering

  {
    \caption{
      Classical features leveraged to calculate baseline classification
      accuracy.
    }\label{tab:classic_features}
  }
  {
    \resizebox{\linewidth}{!}{
    \begin{tabular}{| c |}
      \hline
      \textbf{Feature} \\
      \hline
      \hline
      Median of breathing cycle amplitudes \\
      Interquartile range of breathing cycle amplitudes \\
      Median of breathing cycle widths \\
      Interquartile range of breathing cycle widths \\
      Median of breathing cycle peaks \\
      Interquartile range of breathing cycle peaks \\
      Median of breathing cycle troughs \\
      Interquartile range of breathing cycle troughs \\
      Median of the area between signal and baseline during inhalation (MAI) \\
      Median of the area between signal and baseline during exhalation (MAE) \\
      Ratio of MAI to MAE (MAI / MAE) \\
      \hline
    \end{tabular}
    }
  }
  \vspace{-2em}
\end{table}

\subsection{Topological Data Analysis}\label{subsec:tda}

We use TDA methods to analyze the time-series signal from the airflow
sensor included in our dataset. Interested readers can refer to
\citet{edelsbrunnerComputationalTopologyIntroduction2022} for a full background
on computational topology.

We require a \emph{filtration} of our data, which allows us to summarize the
topological features of our data at different scales using a PD. A filtration is
a sequence of nested subsets of data, formed by varying a \emph{filtration
parameter}. Each filtration parameter corresponds to a threshold value for a
chosen function on the data. The chosen function depends on the filtration type.
At each subset identified by varying the filtration parameter, we calculate the
topological features of the data as described by the homology groups of each
subset. Informally, homology groups measure the topological properties of data,
such as the number of connected components, holes, voids, and other
higher-dimensional features. The $n$th homology group is denoted $H_n$. In this
work, we focus on the 0th and 1st homology groups, and use two types of
filtrations: Vietoris-Rips filtrations, and sublevel set
filtrations~\citep{chazalIntroductionTopologicalData2021c}.


\paragraph{Vietoris-Rips Filtration}\label{subsec:rips_complex}

\begin{figure*}[htbp]
  \label{fig:filtrations}
  \centering
  \caption{\textbf{(a)} Example Rips filtration. As $\epsilon$ increases,
    connections between points are made. When $\epsilon = 3$, a single
    connection between two points is made, creating a feature in $H_0$. When
    $\epsilon = 5.25$, a loop is formed, creating a feature in $H_1$. When
    $\epsilon = 8$, we have a fully connected graph and no further topological
    features will be created as $\epsilon \rightarrow \infty$.\textbf{(b)}
    Example sublevel set filtration from a synthetic time-series signal. We show
    an example topological feature and the corresponding point on the
  persistence diagram. The feature is born at $t_1$ and dies at $t_2$, as
indicated by the ($t_1$, $t_2$) point on the persistence diagram}
  {\resizebox{0.95\textwidth}{!}{
    \includegraphics[width=\linewidth]{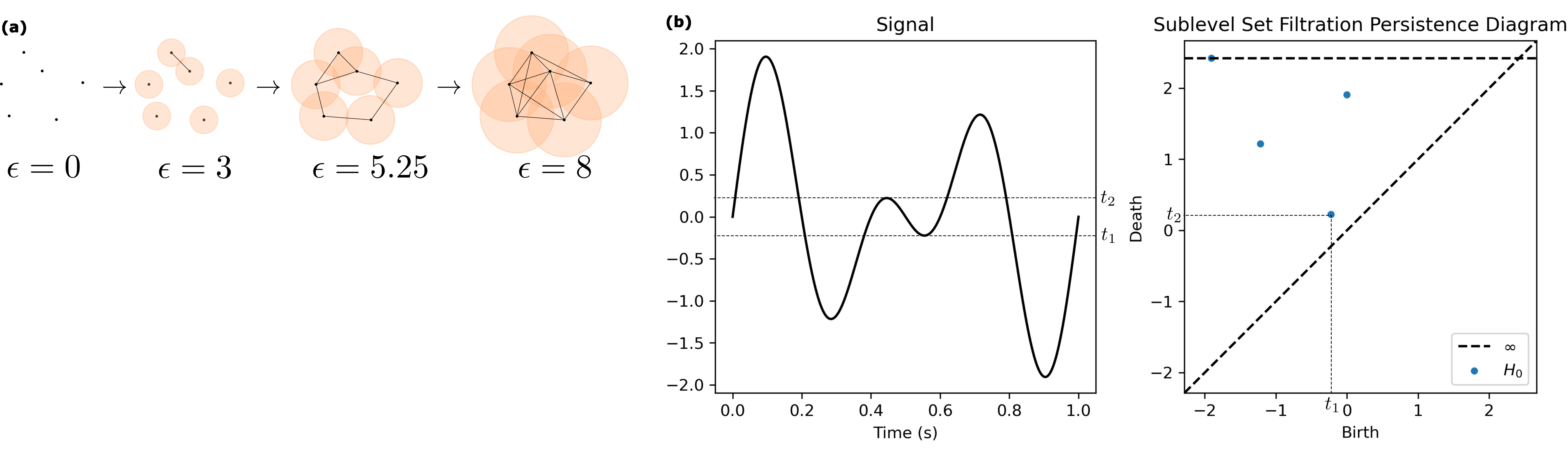}
  }}
  \vspace{-1em}
\end{figure*}

In order to compute a Vietoris-Rips filtration (henceforth referred to as a Rips
filtration), we must first embed our time-series data in a high-dimensional
space. To do this, we use a Takens'
embedding~\citep{takensDetectingStrangeAttractors1981,pereaTopologicalTimeSeries2018}.
Takens' theorem states that, given reasonable assumptions, we can recover the
underlying phase space of a system based on the observed time series. To
construct a Takens' embedding, we first choose a time delay $\tau \in \N$ and an
embedding dimension $d \in \N$. Given a time series $f = \{f_1, \cdots, f_N\}$,
we construct a $d$-dimensional vector $x_1 = \{f_1, f_{1+\tau}, \cdots f_{1 +
(d-1)\tau}\}$ associated with $f_1$. We repeat this process until we have $N - (d-1)\tau$ vectors $X =
\{x_1, \cdots, x_{N - (d-1)\tau}\}$. This represents our high-dimensional
vectors for our Rips filtration. In this work, we set $\tau$ to be the
equivalent number of samples of one second (256 samples for 256 Hz), and $d$ as
3.

We can now construct our Rips complex from our Takens embedding. The Rips
complex $\mathcal{R}$ of $X$ and distance $\epsilon$ is given by all data points
and all connections between data points of $X$ with distance less than or equal
to $\epsilon$. Once we have constructed a Rips filtration by varying the
filtration parameter $\epsilon$ from 0 to infinity, we construct a PD to
describe the topological features of the constructed Rips filtration. The PD
records the ``birth'' and ``death'' times of topological features identified
from our Rips filtration. When a topological feature (a connection between two
points for $H_0$ or a loop for $H_1$) forms when the ball of size $\epsilon$
around one point touches the ball of size $\epsilon$ around another point, we
record the $\epsilon$ value at which it is created as the birth value. When the
feature is destroyed or superseded by another topological feature, we record the
$\epsilon$ value as the death value. We calculate the Rips filtration of the
epoch of interest and 5 previous epochs (180 second of data) on our airflow
signal. If there are any apnea, hypoponea, or other oxygen desaturation events
within 5 previous epochs of interest, we do not use the current target
epoch as part of our dataset.


\paragraph{Sublevel Set Filtration}

Sublevel set filtrations are useful for analyzing the morphology of a given
signal. We construct a sublevel set filtration from a signal $f$ indexed by
$t \in \R$. The sublevel set of $f$ at $t$ is defined as: $f_t^{-} =
\{x | f(x) \leq t\}$. If $t_1 \leq t_2$, then $f_{t_1}^{-} \subseteq
f_{t_2}^{-}$. To construct a persistence diagram from the sublevel set
filtration, similarly to the Rips filtration, we record the $t$ values when
topological a topological feature is created as the birth value and the $t$
value at which the topological feature is destroyed as the death value. Since
sublevel set filtrations are 1-dimensional objects, we only consider $H_0$ for
the sublevel set filtration~\citep{carlssonTopologicalDataAnalysis2021a}. We
calculate the sublevel set filtration of the current epoch and the 5 previous
epochs (180 second windows) of our airflow signal and the current epoch plus the
5 previous epochs of our IRR signal.

\paragraph{Persistence Curves}

We now focus on PCs, a method for PD featurization. Let $D$
denote a PD, a multi-set of pairs of real numbers with a multiplicity function
$m$: $D = \{(b_i, d_i) | b_i \leq d_i, m(b_i, d_i) < \infty\}_{i \in \N}$ The
multiplicity function $m$ allows for the case when multiple topological features
have the same birth or death time. We can now define the PC on a
PD~\citep{chungPersistenceCurvesCanonical2021}.

\begin{definition}[Persistence Curve]\label{eq:pc}

  Let $\psi(b, d, x): \R^3 \rightarrow \R$ with $\psi(b, b, x) = 0$. The
  persistence curve of $D$ with respect to $\psi(b_i, d_i, x)$ is denoted by $P(D,
  \psi)$ and is defined as:

  \[
    P(D, \psi)(x) = \sum\limits_{(b_i, d_i) \in D}\psi(b_i, d_i) \chi_{[b_i, d_i)} (x)
  \]

\end{definition}

\noindent where $\chi_{[b_i, d_i)}$ is the indicator function for the half-open
interval $[b_i, d_i)$. If $\psi$ is known, we denote the PC as
$P(D)$ and omit the $\psi$. In this work, we focus on the lifespan entropy
function as the $\psi$ function~\citep{chungPersistenceCurvesCanonical2021}. Let
$L = \sum_{(b_i, d_i) \in D} (d_i - b_i)$ for a PD $D$. In this case, $\psi(b, d) =
-\frac{d-b}{L}\log(\frac{d-b}{L})$. We
define the lifespan entropy PC as $P(D)(x) = -\sum_{(b_i, d_i)
\in D} \frac{d_i - b_i}{L}\log\left(\frac{d_i - b_i}{L}\right) \chi_{[b_i,
d_i)}(x)$ for $x \in \R$. The
PC $P(D)$ is typically a step function on a grid defined by the
$(b_i, d_i)$ values included in the diagram $D$. However, $P(D)$ can have a
varying number of steps depending on the number of points in the diagram. A
typical method to handle this varying number of steps would be to discretize the
grid on which the PC is defined and sample the PC at those points. This presents
several problems. The number of points in $D$ can be large, leading to a large
number of calculations required to sample the PC. Therefore, recent work has
attempted to approximate PCs using Hermite functions.

\begin{figure*}[htbp]
  \caption{Example persistence diagrams and approximations for associated
    PCs using HEPC, SP-FAPC and AP-FAPC methods. SP-FAPC provides
    slightly better fits since we can control the domain and set it to
    reasonable values. In contrast, the AP-FAPC method using 15 coefficients is
    able to fit the PCs very well at varying scales. However,
    AP-FAPC uses different bases for each fit, so the bases on one diagram are
  different from the bases used on a different diagram.}\label{fig:approx_fits}
  \includegraphics[width=\linewidth]{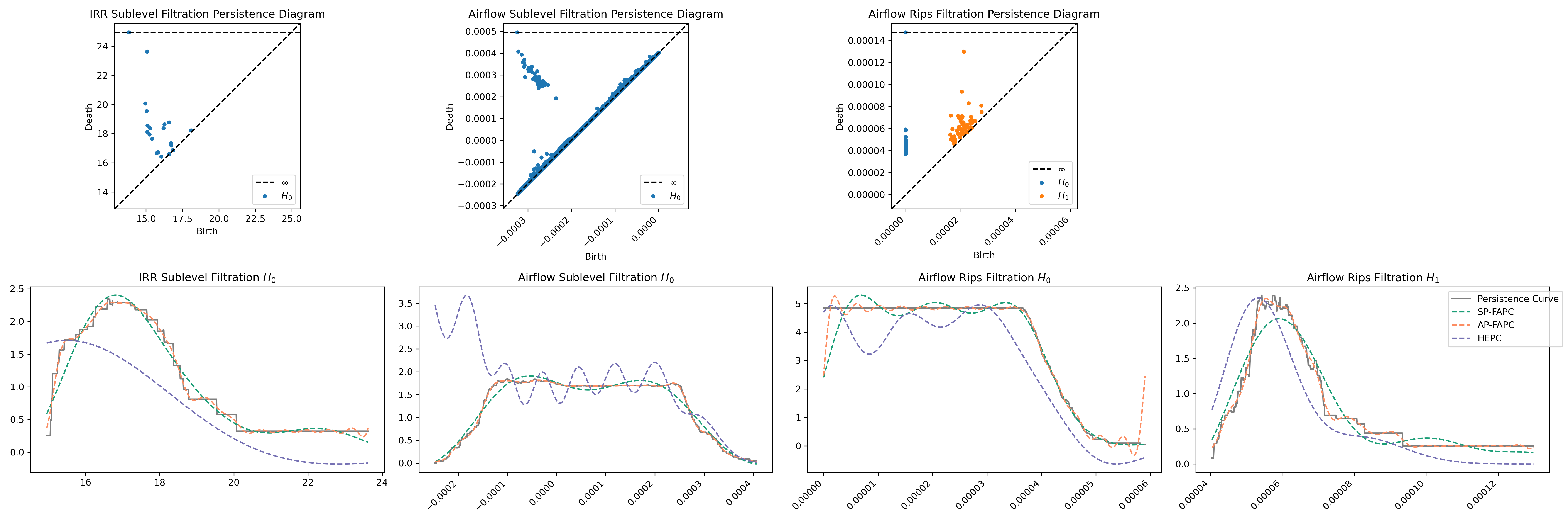}
  \vspace{-3em}
\end{figure*}

\subsubsection{Hermite Expansion of Persistence Curves}\label{subsubsec:hepc}

%

%

We can define the Hermite function $h_n(x)$ using $F_n (x) = (-1)^n e^{x^2}
\frac{d^n}{dx^n}e^{-x^2}$, the physicist's Hermite polynomial: $h_n (x) = c_n
F_n(x) \phi(x)$, where $c_n = (2^n n! \sqrt{\pi})^{-1/2}$ and $\phi(x) =
\frac{1}{\sqrt{2 \pi}} e^{-x^2/2}$, the probability density function of the
standard normal distribution. $\{h_n (x)\}$ forms an orthogonal basis on
$L^2(\R, e^{-x^2}dx)$, the square integrable functions defined on the
real numbers. Therefore, when the PC is in $L^2(\R, e^{-x^2}dx)$, we can uniquely
approximate PCs as $P(D)(x) = \sum_{i=0}^\infty \alpha_i h_i(x)$ for some
coefficients $\alpha_i \in \R$. We can derive the following recursive formula
for $\alpha_i$~\citep{chungStableTopologicalFeature2022}.

\begin{align*}
  \alpha_0 &= \sum\limits_{(b_i, d_i) \in D} \sqrt{2}\pi^{1/4}\psi(b_i,
  d_i)\left[\Phi(d_i) - \Phi(b_i)\right] \\
  \alpha_1 &= \sum\limits_{(b_i, d_i) \in D} 2 \pi^{1/4}\psi(b_i,
  d_i)\left[\phi(b_i) - \phi(d_i)\right] \\
  \alpha_{n+1} &= \frac{\sqrt{2}}{\sqrt{n+1}} \left[\sum\limits_{(b_i, d_i) \in
  D} \psi(b_i, d_i)(\psi_n(b_i) - \psi_n(d_i))\right] \\
               & \qquad\qquad + \frac{n}{\sqrt{n(n+1)}}\alpha_{n-1}
\end{align*}

\noindent where $\Phi$ is the cumulative density fuction of the standard normal
distribution.

In this work, we compute the first 15 coefficients
and place them in a vector $[\alpha_0, \cdots, \alpha_{14}]^\top$. This is to
match the prior work of \citet{chungTopologicalDataAnalysis2024} using HEPC
methods for TDA-based airflow sleep staging.

In our dataset, our PCs exist on grids at very small scales, as indicated in the
$d_{\text{max}}$ column of Table~\ref{tab:residuals}. $d_\text{max}$ is the
maximum non-infinity death value for an individual PD, and
indicates the scale of the PD obtained from a particular
filtration. The default small $d_\text{max}$ values lead to poor Hermite
expansion fits, as higher order Hermite functions are required to fit data on
small scales. Therefore, we scale our PCs by scalar constants specific to each
signal and filtration in
order to promote reasonable Hermite function fits. Full details of scalar
constant calculation can be found in Appendix~\ref{apd:fit_constants}.

\subsubsection{Fourier Approximation of Persistence Curves}\label{subsubsec:fapc}

Hermite functions are able to reasonably fit the general trend of
PCs; however, many of the finer details are lost by the Hermite
function fit. We now present our novel contribution on approximating persistence
curves on arbitrarily small or large intervals, Fourier Approximation of
Persistence Curves (FAPC). Consider the set of functions $\{e_n = e^{i n x}\}$,
where $i$ is the imaginary number. This set forms an orthonormal basis for the
space $L^2([0, 2\pi])$. We can dilate this basis to the space $L^2([\dmin,
\dmax])$ by using the set of functions $\{e_n = e^{i 2 \pi x / (\dmax -
\dmin)}\}$. We can approximate our PC, denoted $P(D)$, using a Fourier
series: $P(D) = \sum_{n=-\infty}^\infty \beta_n e_n$. We additionally derive a
closed-form solution for the Fourier coefficients in
Appendix~\ref{apd:fapc_closed_form} which does not require sampling of the
domain of the PC. However, we still must choose two parameters, $\dmin$ and $\dmax$.

Our first method for handling $\dmin$ and $\dmax$ is entirely based on the
specific PC being approximated. We call this method ``Arbitrary Period FAPC
(AP-FAPC),'' and choose our $\dmin$ and $\dmax$ values based on the interval in
the domain where the PC has nonzero value. This method allows for very low
residual values; however, since the bases $\{e_n\}$ change when the $\dmin$ and
$\dmax$ values change, we cannot compare the coefficients between two different
PCs, as the basis of each approximation is different. This presents a problem,
since two different PCs can have the same AP-FAPC coefficients since scale
information is not captured by the AP-FAPC method. Therefore, we lose
information about the specific domain of the PC, but gain accurate approximation
of the PC morphology.

The second method we propose is based on set $\dmin$ and $\dmax$ values,
allowing consistent basis functions across all PC fits. We set the $\dmin$ and
$\dmax$ values to predetermined numbers specific to each signal type and
filtration type (e.g., airflow Rips filtration has one pair of $\dmin$ and
$\dmax$ values, while airflow sublevel set filtration has different $\dmin$ and
$\dmax$ values), then calculate the coefficients of PCs based on these preset
$\dmin$ and $\dmax$ values. We call this method ``Set Period FAPC (SP-FAPC).''
We choose the $\dmin$ and $\dmax$ values based on reasonable clinical values for
each sensor type and filtration method. Specifics on the $\dmin$ and $\dmax$
values for SP-FAPC can be found in Appendix~\ref{apd:fit_constants}. In this
method, we have consistent bases $\{e_n\}$ between PCs. However, the nonzero
domain can vary significantly between different PCs. If a PC has a much smaller
domain than the selected $\dmin$ and $\dmax$ values, the PC is relatively
high-frequency, and we lose information about the specific morphology of the PC
when use use a limited number of coefficients.

In this work, we compute the first 15 coefficients $[\beta_0, \cdots,
\beta_{14}]$ and place them in a vector to match the number of coefficients
computed in the HEPC feature as described in Section~\ref{subsubsec:hepc}.
However, the Fourier approximation leads to a vector in $\C^{15}$, where $\C$
denotes the complex field. Therefore, we take the real portion of each element
and the imaginary portion of each element and concatenate them into a vector in
$\R^{30}$.

For AP-FAPC, while each approximation runs over an arbitrary interval $A$ based
on the specific PC being approximated, the approximations are
much closer to the true shape of the PC and capture the specifics
of the signal morphology, even if the scale is small or large as indicated in
Table~\ref{tab:residuals}. For SP-FAPC we observe higher residuals. There is
significant variation in the domain of PCs; some PCs have a domain which is only
a small subset of the reasonable clinical values for that PC. Therefore, due to
the limited 15-coefficient approximation, many of the high-frequency contents of
the PC are lost, as in HEPC approximations. The advantage of SP-FAPC over
AP-FAPC is that it maintains consistent basis functions $\{e_n\}$ for each PC,
leading to approximation coefficients which are consistent across different
domains.

\subsection{Machine Learning Methods}

\paragraph{Features}

We use several feature sets in our evaluation of the FAPC method and our broader
evaluation of TDA for airflow-based sleep staging. We first include non-TDA
baseline features described in Table~\ref{tab:classic_features} to create a
baseline for model performance. We compute HEPC and FAPC approximations of the
$H_0$ persistence curve from the Rips filtration of the airflow signal, the
$H_1$ persistence curve from the Rips filtration of the airflow signal, the
$H_0$ diagram from the sublevel set filtration of the airflow signal, and the
$H_0$ diagram from the sublevel set filtration of the IRR signal.
Prior to training we perform Z-normalization on all features in all epochs of
the training set, then Z-normalize features in the testing set based on the mean
and standard deviation of the training set.

\paragraph{Model Training and Testing}

We use an XGBoost model for this work~\citep{chenXGBoostScalableTree2016}.
Due to the class imbalance in our Wake/NREM/REM labels, we weight the XGBoost
loss function based on the inverse of the class frequencies in the training set.
We include specific XGBoost parameters in Appendix~\ref{apd:xgboost_params}. To
appropriately characterize the performance of our model, we leverage a
per-subject 5-fold cross validation, i.e. all data from an individual subject
will be in the test or train set. Since our population for this dataset is a
pediatric population with a variety of age and sex combinations, we age and sex
match our cross validation splits such that the same proportion of each age/sex
group is in the train and test sets. When testing statistical significance, we
use a paired $t$-test with a significance level of $0.05$.

%
%


\section{Results and Discussion} \label{sec:results}

\begin{figure}[htbp]
  \vspace{-2em}
  \caption{Statistical significance of differences of test balanced accuracy on
  different feature sets calculated with a paired t-test. * indicates $p <
  0.05$, ** indicates $p<0.005$, and *** indicates $p < 0.0005$.}
  \label{fig:stat_sig}
  \includegraphics[width=\linewidth]{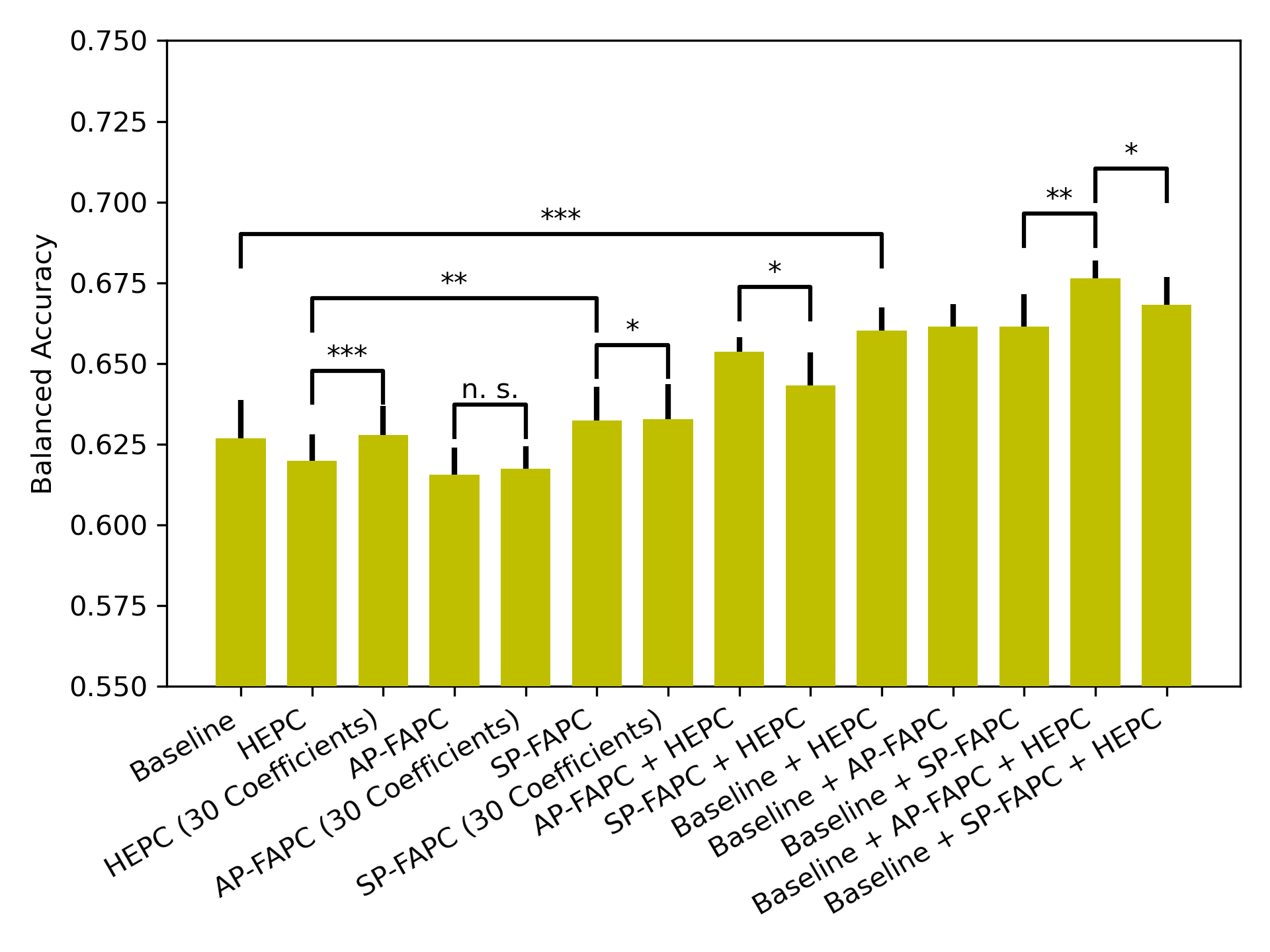}
  \vspace{-2em}
\end{figure}

\begin{table*}[htbp]
  \vspace{-1em}
  \centering
  {
    \caption{
      Results of XGBoost model on various feature types. We present Train
      and Test Balanced Accuracy (BA) across the Wake/NREM/REM classes averaged
      across our 5-fold cross validation, as well as the individual Wake/NREM/REM
      test set class accuracies averaged across our 5-fold cross validation.
    } \label{tab:results}
  }
  {
    \resizebox{0.7\textwidth}{!}{
      \begin{tabular}{| c || c | c | c | c | c | c |}
        \hline
        Feature Type & Train BA & Test BA & Cohen's $\kappa$ &  Wake & REM & NREM \\
        \hline
        \hline
        Random Prediction & 0.376 & 0.333 & -0.000 & 0.341 & 0.305 & 0.353 \\
        \hline
        Baseline & 0.650 & 0.627 & 0.335 & 0.598 & 0.683 & 0.599 \\
        \hline
        HEPC & 0.636 & 0.620 & 0.286 & 0.656 & 0.610 & 0.593 \\
        \hline
        HEPC (30 Coefficients) & 0.645 & 0.628 & 0.292 & 0.670 & 0.610 & 0.603 \\
        \hline
        AP-FAPC & 0.629 & 0.616 & 0.256 & 0.736 & 0.547 & 0.564 \\
        \hline
        AP-FAPC (30 Coefficients) & 0.630 & 0.617 & 0.257 & 0.738 & 0.547 & 0.566 \\
        \hline
        SP-FAPC & 0.650 & 0.632 & 0.299 & 0.664 & 0.611 & 0.622 \\
        \hline
        SP-FAPC (30 Coefficients) & 0.651 & 0.633 & 0.300 & 0.663 & 0.613 & 0.622 \\
        \hline
        AP-FAPC + HEPC & 0.670 & 0.654 & 0.323 & 0.710 & 0.633 & 0.618 \\
        \hline
        SP-FAPC + HEPC & 0.662 & 0.643 & 0.305 & 0.679 & 0.611 & 0.640 \\
        \hline
        Baseline + HEPC & 0.681 & 0.660 & 0.350 & 0.676 & 0.670 & 0.634 \\
        \hline
        Baseline + AP-FAPC & 0.679 & 0.661 & 0.340 & 0.722 & 0.654 & 0.608 \\
        \hline
        Baseline + SP-FAPC & 0.681 & 0.661 & 0.347 & 0.677 & 0.660 & 0.647 \\
        \hline
        Baseline + AP-FAPC + HEPC & 0.697 & 0.676 & 0.359 & 0.719 & 0.665 & 0.645 \\
        \hline
        Baseline + SP-FAPC + HEPC & 0.689 & 0.668 & 0.352 & 0.686 & 0.661 & 0.658 \\
        \hline
      \end{tabular}
    }
  }

  \centering
  {
    \caption{
        Average Residual for HEPC and FAPC approximations calculated on
        persistence diagrams from different Homology groups from the dataset. We
        additionally include the maximum non-infinity death value found in the dataset
        for the particular filtration, denoted $d_{\text{max}}$. $d_\text{max}$ indicates the
        scale of data for a particular filtration.
    } \label{tab:residuals}
  }
  {
    \resizebox{0.8\textwidth}{!}{
      \begin{tabular}{| c || c | c | c | c | c | c | c |}
        \hline
        Residual & HEPC & SP-FAPC  & AP-FAPC & Minimum $\dmin$ & Median $\dmin$ &  Median $\dmax$ & Maximum $\dmax$ \\
        \hline
        \hline
        $H_0$ Rips Filtration of Airflow Signal & 6540.471 & 12138.079 & 80.460 & 0.0 & 0.0 & $1.304 \times 10^{-4}$ & $3.683 \times 10^{-3}$\\
        \hline
        $H_1$ Rips Filtration of Airflow Signal & 1337.812 & 282.142 & 2.412 & $5.758 \times 10^{-7}$ & $8.250 \times 10^{-5}$ & $2.867 \times 10^{-4}$ & $4.183 \times 10^{-3}$\\
        \hline
        $H_0$ Sublevel Filtration of Airflow Signal & 2359.405 & 1021.375 & 4.921 & $-2.000 \times 10^{-3}$ & $-6.366 \times 10^{-4}$ & $6.668 \times 10^{-4}$ & $2.000 \times 10^{-3}$ \\
        \hline
        $H_0$ Sublevel Filtration of IRR Signal & 1007.144 & 312.940 & 2.815 & -10.295 & 11.973 & 32.762 & 396.521 \\
        \hline
      \end{tabular}
    }
  }
  \vspace{-1em}
\end{table*}

In this study, we propose a novel method for approximating PCs,
AP-FAPC and SP-FAPC, which outperforms the existing state-of-the-art for
approximating PCs, HEPC. AP-FAPC is able to much more accurately
approximate PCs than SP-FAPC or HEPC approximations. Our results based
on the residual of our fitting methods in Table~\ref{tab:residuals} indicates
that AP-FAPC and SP-FAPC can consistently fit the various signals involved in
respiration-based sleep staging, while HEPC methods are only able to effectively
approximate the data when the scale is within a specific range. HEPC is only
able to achieve a lower residual SP-FAPC on $H_0$ Rips filtrations of airflow
signals; in all other cases, SP-FAPC and AP-FAPC achieve lower residuals than
HEPC.

Much of the difficulty with featurizing PDs, especially in medical problems,
comes from the significant variability between PDs generated by different human
subjects. This is exacerbated by our use of a pediatric cohort, which leads to
greater variability due to child development. Despite this significant
variability, we are able to achieve a test balanced accuracy of 67.6\% and a
Cohen's $\kappa$ of 0.360 using a combination of baseline, AP-FAPC, and HEPC
features with an XGBoost model. We include a test set confusion matrix averaged
over all 5 folds in Appendix~\ref{apd:confusion_matrix}.

Our results based on training an XGBoost model for airflow-based sleep staging
indicate that individually, SP-FAPC approximations significantly outperform HEPC
approximations as shown in Table~\ref{tab:results} ($p = 0.001$ based on a
paired $t$-test as shown in \figureref{fig:stat_sig}), while AP-FAPC methods
perform worse than HEPC or SP-FAPC methods. Since AP-FAPC uses varying basis
functions, no information on the domain of the PC is retained;
the AP-FAPC coefficients only contain information about the specific morphology
of the data. Compared to baseline, HEPC and AP-FAPC perform worse than the
baseline, while SP-FAPC does slightly outperform the baseline features based on
test balanced accuracy, but this difference is not statistically significant.

While AP-FAPC approximations perform poorly on their own, they perform
significantly better when combined with HEPC coefficients, achieving a test
balanced accuracy of 65.4\%. SP-FAPC coefficients combined with HEPC
coefficients only achieve a test balanced accuracy of 64.3\%. This is a
statistically significant difference. Therefore, combining the HEPC
approximation, which attempts to approximate the general trend of PCs while
maintaining information about the underlying PC domain, with the specific
morphology of the PC encoded by the AP-FAPC coefficients leads to better
performance than combining HEPC coefficients with SP-FAPC coefficients
calculated on a set domain. Based on these results, we conclude that the
specific morphology of the PC is highly important for downstream classification
accuracy using machine learning models, not just the general trend of the PC. We
furthermore note that the difference in performance is not due to the extra
features passed to the model alone. When trained with 30 coefficients as opposed
to 15, HEPC, and SP-FAPC models perform only slightly better than 15-coefficient
models. AP-FAPC performs almost exactly the same as more coefficients are added,
as the approximation with 15 coefficients is quite good due to the basis
functions being specific to the PC being approximated.

Models trained on baseline and SP-FAPC, baseline and AP-FAPC, or baseline and
HEPC features provide a statistically significant improvement over models
trained on only baseline ($p < 0.0005$), as shown in \figureref{fig:stat_sig}.
We find that FAPC and HEPC methods provide complimentary information to the
classic features, indicating that TDA-based methods provide additional
information which is not included in our non-TDA baseline feature set. Our best
test accuracy combining baseline AP-FAPC, and HEPC features achieves an overall
test accuracy of 67.6\%. This is not high enough for clinical use, but does
represent a step forward in eliminating the need for EEG data in sleep staging.
We furthermore note that since we use a dataset of pediatric subjects, our
model is able to achieve 67.6\% accuracy across a wide range of age groups from
ages 2-18, indicating the power of this method for sleep staging. Further work
will involve testing this method on an adult population.


There are several limitations to this study. First, computing TDA features is
computationally intensive, and requires significant computation time. Recent
advancements in computation of both Rips and sublevel set filtrations have
decreased the amount of time required to compute filtrations; however, when
there are a large number of samples due to high sampling rates or long
time-series signals, computation can take a significant amount of time,
especially with a large number of subjects. Future work may include studying
alternate basis functions such as Haar wavelets or other wavelet functions which
are able to better approximate the specific morphology of PCs at varying scales,
while maintaining information about the domain of the persistence function.
Additional future work includes testing the FAPC methods on other health related
tasks as well as on non-health-related tasks.

\section{Conclusion}\label{sec:conclusion}

In this work we developed Fourier approximations of PCs, a method
for approximating PCs. We applied this method to a large dataset
of pediatric subjects, and compared this method to the existing Hermite
approximation of PCs as well as baseline methods. We find that
Fourier approximations outperform Hermite approximations on airflow-based sleep
staging tasks.

%

\bibliography{smanjunath_ml4h_tda_sleep_staging}

\begin{thebibliography}{31}
\providecommand{\natexlab}[1]{#1}
\providecommand{\url}[1]{\texttt{#1}}
\expandafter\ifx\csname urlstyle\endcsname\relax
  \providecommand{\doi}[1]{doi: #1}\else
  \providecommand{\doi}{doi: \begingroup \urlstyle{rm}\Url}\fi

\bibitem[Adams et~al.(2017)Adams, Emerson, Kirby, Neville, Peterson, Shipman, Chepushtanova, Hanson, Motta, and Ziegelmeier]{adamsPersistenceImagesStable2017}
Henry Adams, Tegan Emerson, Michael Kirby, Rachel Neville, Chris Peterson, Patrick Shipman, Sofya Chepushtanova, Eric Hanson, Francis Motta, and Lori Ziegelmeier.
\newblock Persistence {{Images}}: {{A Stable Vector Representation}} of {{Persistent Homology}}.
\newblock \emph{Journal of Machine Learning Research}, 18\penalty0 (8):\penalty0 1--35, 2017.
\newblock ISSN 1533-7928.

\bibitem[Bakker et~al.(2021)Bakker, Ross, Vasko, Cerny, Fonseca, Jasko, Shaw, White, and Anderer]{bakkerEstimatingSleepStages2021}
Jessie~P. Bakker, Marco Ross, Ray Vasko, Andreas Cerny, Pedro Fonseca, Jeff Jasko, Edmund Shaw, David~P. White, and Peter Anderer.
\newblock Estimating sleep stages using cardiorespiratory signals: Validation of a novel algorithm across a wide range of sleep-disordered breathing severity.
\newblock \emph{Journal of Clinical Sleep Medicine : JCSM : Official Publication of the American Academy of Sleep Medicine}, 17\penalty0 (7):\penalty0 1343--1354, July 2021.
\newblock ISSN 1550-9389.
\newblock \doi{10.5664/jcsm.9192}.

\bibitem[Barnes et~al.(2021)Barnes, Polanco, and Perea]{barnesComparativeStudyMachine2021a}
Danielle Barnes, Luis Polanco, and Jose~A. Perea.
\newblock A {{Comparative Study}} of {{Machine Learning Methods}} for {{Persistence Diagrams}}.
\newblock \emph{Frontiers in Artificial Intelligence}, 4, July 2021.
\newblock ISSN 2624-8212.
\newblock \doi{10.3389/frai.2021.681174}.

\bibitem[Birrenkott()]{birrenkottRespiratoryQualityIndex}
Drew Birrenkott.
\newblock Respiratory {{Quality Index Design}} and {{Validation}} for {{ECG}} and {{PPG Derived Respiratory Data}}.

\bibitem[Bubenik()]{bubenikStatisticalTopologicalData}
Peter Bubenik.
\newblock Statistical {{Topological Data Analysis}} using {{Persistence Landscapes}}.

\bibitem[Campbell(2009)]{campbellEEGRecordingAnalysis2009}
Ian~G. Campbell.
\newblock {{EEG Recording}} and {{Analysis}} for {{Sleep Research}}.
\newblock \emph{Current protocols in neuroscience / editorial board, Jacqueline N. Crawley ... [et al.]}, CHAPTER:\penalty0 Unit10.2, October 2009.
\newblock ISSN 1934-8584.
\newblock \doi{10.1002/0471142301.ns1002s49}.

\bibitem[Carlsson and {Vejdemo-Johansson}(2021)]{carlssonTopologicalDataAnalysis2021a}
Gunnar Carlsson and Mikael {Vejdemo-Johansson}.
\newblock \emph{Topological {{Data Analysis}} with {{Applications}}}.
\newblock Cambridge University Press, Cambridge, 2021.
\newblock ISBN 978-1-108-83865-8.
\newblock \doi{10.1017/9781108975704}.

\bibitem[Chazal and Michel(2021)]{chazalIntroductionTopologicalData2021c}
Fr{\'e}d{\'e}ric Chazal and Bertrand Michel.
\newblock An {{Introduction}} to {{Topological Data Analysis}}: {{Fundamental}} and {{Practical Aspects}} for {{Data Scientists}}.
\newblock \emph{Frontiers in Artificial Intelligence}, 4, September 2021.
\newblock ISSN 2624-8212.
\newblock \doi{10.3389/frai.2021.667963}.

\bibitem[Chen et~al.(2023)Chen, Wu, and Chen]{chenQualityAwareSleep2023}
Hsin-Yu Chen, Hau-Tieng Wu, and Cheng-Yao Chen.
\newblock Quality {{Aware Sleep Stage Classification}} over {{RIP Signals}} with {{Persistence Diagrams}}.
\newblock In \emph{2023 {{IEEE}} 19th {{International Conference}} on {{Body Sensor Networks}} ({{BSN}})}, pages 1--4, October 2023.
\newblock \doi{10.1109/BSN58485.2023.10331130}.

\bibitem[Chen and Guestrin(2016)]{chenXGBoostScalableTree2016}
Tianqi Chen and Carlos Guestrin.
\newblock {{XGBoost}}: {{A Scalable Tree Boosting System}}.
\newblock In \emph{Proceedings of the 22nd {{ACM SIGKDD International Conference}} on {{Knowledge Discovery}} and {{Data Mining}}}, pages 785--794, August 2016.
\newblock \doi{10.1145/2939672.2939785}.

\bibitem[Chung(2022)]{chungStableTopologicalFeature2022}
Yu-Min Chung.
\newblock Stable {{Topological Feature Vectors}} via {{Hermite Function Expansion}} on {{Persistence Curves}}.
\newblock In \emph{2022 {{IEEE International Conference}} on {{Big Data}} ({{Big Data}})}, pages 5434--5443, December 2022.
\newblock \doi{10.1109/BigData55660.2022.10020423}.

\bibitem[Chung and Lawson(2021)]{chungPersistenceCurvesCanonical2021}
Yu-Min Chung and Austin Lawson.
\newblock Persistence {{Curves}}: {{A}} canonical framework for summarizing persistence diagrams, August 2021.

\bibitem[Chung et~al.(2024)Chung, Huang, and Wu]{chungTopologicalDataAnalysis2024}
Yu-Min Chung, Whitney~K. Huang, and Hau-Tieng Wu.
\newblock Topological data analysis assisted automated sleep stage scoring using airflow signals.
\newblock \emph{Biomedical Signal Processing and Control}, 89:\penalty0 105760, March 2024.
\newblock ISSN 17468094.
\newblock \doi{10.1016/j.bspc.2023.105760}.

\bibitem[Edelsbrunner and Harer(2022)]{edelsbrunnerComputationalTopologyIntroduction2022}
Herbert Edelsbrunner and John~L. Harer.
\newblock \emph{Computational {{Topology}}: {{An Introduction}}}.
\newblock American Mathematical Society, January 2022.
\newblock ISBN 978-1-4704-6769-2.

\bibitem[Gerstenslager and Slowik(2024)]{gerstenslagerSleepStudy2024}
Brian Gerstenslager and Jennifer~M. Slowik.
\newblock Sleep {{Study}}.
\newblock In \emph{{{StatPearls}}}. StatPearls Publishing, Treasure Island (FL), 2024.

\bibitem[Gouthro and Slowik(2024)]{gouthroPediatricObstructiveSleep2024}
Kathryn Gouthro and Jennifer~M. Slowik.
\newblock Pediatric {{Obstructive Sleep Apnea}}.
\newblock In \emph{{{StatPearls}}}. StatPearls Publishing, Treasure Island (FL), 2024.

\bibitem[Karna et~al.(2024)Karna, Sankari, and Tatikonda]{karnaSleepDisorder2024}
Bibek Karna, Abdulghani Sankari, and Geethika Tatikonda.
\newblock Sleep {{Disorder}}.
\newblock In \emph{{{StatPearls}}}. StatPearls Publishing, Treasure Island (FL), 2024.

\bibitem[Lee et~al.(2022)Lee, Li, DeForte, Splaingard, Huang, Chi, and Linwood]{leeLargeCollectionRealworld2022c}
Harlin Lee, Boyue Li, Shelly DeForte, Mark~L. Splaingard, Yungui Huang, Yuejie Chi, and Simon~L. Linwood.
\newblock A large collection of real-world pediatric sleep studies.
\newblock \emph{Scientific Data}, 9\penalty0 (1):\penalty0 421, July 2022.
\newblock ISSN 2052-4463.
\newblock \doi{10.1038/s41597-022-01545-6}.

\bibitem[Makowski et~al.(2021)Makowski, Pham, Lau, Brammer, Lespinasse, Pham, Sch{\"o}lzel, and Chen]{makowskiNeuroKit2PythonToolbox2021}
Dominique Makowski, Tam Pham, Zen~J. Lau, Jan~C. Brammer, Fran{\c c}ois Lespinasse, Hung Pham, Christopher Sch{\"o}lzel, and S.~H.~Annabel Chen.
\newblock {{NeuroKit2}}: {{A Python}} toolbox for neurophysiological signal processing.
\newblock \emph{Behavior Research Methods}, 53\penalty0 (4):\penalty0 1689--1696, August 2021.
\newblock ISSN 1554-3528.
\newblock \doi{10.3758/s13428-020-01516-y}.

\bibitem[Manjunath et~al.(2023)Manjunath, Perea, and Sathyanarayana]{manjunathTopologicalDataAnalysis2023b}
Shashank Manjunath, Jose~A. Perea, and Aarti Sathyanarayana.
\newblock Topological {{Data Analysis}} of {{Electroencephalogram Signals}} for {{Pediatric Obstructive Sleep Apnea}}.
\newblock In \emph{2023 45th {{Annual International Conference}} of the {{IEEE Engineering}} in {{Medicine}} \& {{Biology Society}} ({{EMBC}})}, pages 1--4, July 2023.
\newblock \doi{10.1109/EMBC40787.2023.10340674}.

\bibitem[Meltzer et~al.(2010)Meltzer, Johnson, Crosette, Ramos, and Mindell]{meltzerPrevalenceDiagnosedSleep2010}
Lisa~J. Meltzer, Courtney Johnson, Jonathan Crosette, Mark Ramos, and Jodi~A. Mindell.
\newblock Prevalence of {{Diagnosed Sleep Disorders}} in {{Pediatric Primary Care Practices}}.
\newblock \emph{Pediatrics}, 125\penalty0 (6):\penalty0 e1410--e1418, June 2010.
\newblock ISSN 0031-4005.
\newblock \doi{10.1542/peds.2009-2725}.

\bibitem[Perea(2018)]{pereaTopologicalTimeSeries2018}
Jose~A. Perea.
\newblock Topological {{Time Series Analysis}}, November 2018.

\bibitem[Perea et~al.(2023)Perea, Munch, and Khasawneh]{pereaApproximatingContinuousFunctions2023}
Jose~A. Perea, Elizabeth Munch, and Firas~A. Khasawneh.
\newblock Approximating {{Continuous Functions}} on {{Persistence Diagrams Using Template Functions}}.
\newblock \emph{Foundations of Computational Mathematics}, 23\penalty0 (4):\penalty0 1215--1272, August 2023.
\newblock ISSN 1615-3375, 1615-3383.
\newblock \doi{10.1007/s10208-022-09567-7}.

\bibitem[Polanco and Perea(2019)]{polancoAdaptiveTemplateSystems2019a}
Luis Polanco and Jose~A. Perea.
\newblock Adaptive {{Template Systems}}: {{Data-Driven Feature Selection}} for {{Learning}} with {{Persistence Diagrams}}.
\newblock In \emph{2019 18th {{IEEE International Conference On Machine Learning And Applications}} ({{ICMLA}})}, pages 1115--1121, December 2019.
\newblock \doi{10.1109/ICMLA.2019.00186}.

\bibitem[Takens(1981)]{takensDetectingStrangeAttractors1981}
Floris Takens.
\newblock Detecting strange attractors in turbulence.
\newblock In David Rand and Lai-Sang Young, editors, \emph{Dynamical {{Systems}} and {{Turbulence}}, {{Warwick}} 1980}, volume 898, pages 366--381. Springer Berlin Heidelberg, Berlin, Heidelberg, 1981.
\newblock ISBN 978-3-540-11171-9 978-3-540-38945-3.
\newblock \doi{10.1007/BFb0091924}.

\bibitem[Tataraidze et~al.(2015)Tataraidze, Anishchenko, Korostovtseva, Kooij, Bochkarev, and Sviryaev]{tataraidzeSleepStageClassification2015}
Alexander Tataraidze, Lesya Anishchenko, Lyudmila Korostovtseva, Bert~Jan Kooij, Mikhail Bochkarev, and Yurii Sviryaev.
\newblock Sleep stage classification based on respiratory signal.
\newblock In \emph{2015 37th {{Annual International Conference}} of the {{IEEE Engineering}} in {{Medicine}} and {{Biology Society}} ({{EMBC}})}, pages 358--361, August 2015.
\newblock \doi{10.1109/EMBC.2015.7318373}.

\bibitem[Tian et~al.(2017)Tian, Hu, Qi, Ye, Che, Ding, and Peng]{tianHierarchicalClassificationMethod2017}
Pan Tian, Jie Hu, Jin Qi, Xian Ye, Datian Che, Ying Ding, and Yinghong Peng.
\newblock A hierarchical classification method for automatic sleep scoring using multiscale entropy features and proportion information of sleep architecture.
\newblock \emph{Biocybernetics and Biomedical Engineering}, 37\penalty0 (2):\penalty0 263--271, January 2017.
\newblock ISSN 0208-5216.
\newblock \doi{10.1016/j.bbe.2017.01.005}.

\bibitem[Xu et~al.(2021)Xu, Drougard, and Roy]{xuTopologicalDataAnalysis2021a}
Xiaoqi Xu, Nicolas Drougard, and Rapha{\"e}lle~N. Roy.
\newblock Topological {{Data Analysis}} as a {{New Tool}} for {{EEG Processing}}.
\newblock \emph{Frontiers in Neuroscience}, 15:\penalty0 761703, November 2021.
\newblock ISSN 1662-4548.
\newblock \doi{10.3389/fnins.2021.761703}.

\bibitem[Zhang et~al.(2018)Zhang, Cui, Mueller, Tao, Kim, Rueschman, Mariani, Mobley, and Redline]{zhangNationalSleepResearch2018c}
Guo-Qiang Zhang, Licong Cui, Remo Mueller, Shiqiang Tao, Matthew Kim, Michael Rueschman, Sara Mariani, Daniel Mobley, and Susan Redline.
\newblock The {{National Sleep Research Resource}}: Towards a sleep data commons.
\newblock \emph{Journal of the American Medical Informatics Association: JAMIA}, 25\penalty0 (10):\penalty0 1351--1358, October 2018.
\newblock ISSN 1527-974X.
\newblock \doi{10.1093/jamia/ocy064}.

\bibitem[Zhang et~al.()Zhang, Zhang, Huang, Lv, and Chen]{zhangReviewAutomatedSleep}
Xiaoli Zhang, Xizhen Zhang, Qiong Huang, Yang Lv, and Fuming Chen.
\newblock A review of automated sleep stage based on {{EEG}} signals.

\bibitem[Zheng et~al.(2023)Zheng, Feng, Li, Liang, Cao, and Ge]{zhengTopologicalDataAnalysis2023}
Jingyi Zheng, Ziqin Feng, Yuexin Li, Fan Liang, Xuan Cao, and Linqiang Ge.
\newblock Topological {{Data Analysis}} for {{Scalp EEG Signal Processing}}.
\newblock In \emph{2023 8th {{International Conference}} on {{Signal}} and {{Image Processing}} ({{ICSIP}})}, pages 549--553, July 2023.
\newblock \doi{10.1109/ICSIP57908.2023.10270899}.

\end{thebibliography}

\appendix

\section{Derivation of Closed-Form Solution for Fourier Approximation of
Persistence Curves}\label{apd:fapc_closed_form}

Consider a persistence diagram $D = \{(b_k, d_k) | b_k \leq d_k, m(b_k, d_k) <
\infty\}_{k \in \N}$, where $b_k \in \R$, $d_k \in \R$, and $m$ is a
multiplicity function. We define the persistence curve on $D$ as

\begin{equation}\label{eq:pd}
  P(D)(x) = \sum\limits_{(b, d) \in D} \psi(b, d) \chi_{[b, d)}(x)
\end{equation}

\noindent where $\chi_{[b, d)}(x)$ is the indicator function for the half-open
interval $[b, d)$. It is important to note that the persistence curve $P(D)(x)$
is a continuous step function, as the indicator function limits the nonzero
values of the persistence curve to the domain from the minimum birth value to
the maximum death value.

The $\psi(b, d)$ function is, in general, left as a choice to the user; the only
constraint is that it can only be a function of $b$ and $d$, and not of $x$. In
this paper, we use the lifespan entropy function as our $\psi$ function:

\begin{align*}
  L &= \sum\limits_{(b, d) \in D} (d - b) \\
  \psi(b, d) &= - \frac{d - b}{L} \log\left(\frac{d - b}{L}\right)
\end{align*}

Our goal in this work is to approximate $PD(x)$ with a Fourier series, as given
in Equation~\eqref{eq:pd_fseries}.

\begin{equation}\label{eq:pd_fseries}
  P(D)(x) = \sum\limits_{n = 0}^\infty \beta_n e_n(x)
\end{equation}

where $e_n(x) = e^{2 \pi i n x / A}$, $i$ denotes the imaginary number, and $A =
d_\text{max} - d_\text{min}$, the period of the persistence diagram. Note
that $\{e_n\}$ forms an orthonormal basis of $L^2([d_\text{min}, d_\text{max}])$. In this case, we set
the period as $A = d_{\text{max}} - d_\text{min}$, the difference maximum
non-infinity death value and the minimum birth value in our persistence
diagram $D$. Therefore, $P(D)$ is entirely contained within our domain
$[d_\text{min}, d_\text{max}]$.

Since $\{e_n\}$ forms an orthonormal basis for our domain, we can calculate the
coefficients $\beta_n$ as follows.

\[
  \beta_n = \langle P(D)(x), e_n \rangle = \frac{2}{A} \int_{\dmin}^{\dmax} PD(x) e^{2
  \pi i n x / A}
\]

We first derive $\beta_0$.

\begin{align*}
  \beta_0 &= \frac{2}{A} \int_{\dmin}^{\dmax} \sum\limits_{(b, d) \in D} \psi(b, d) \chi_{[b, d)} (x)
  e^0 dx \\
  \beta_0 &= \frac{2}{A} \sum\limits_{(b, d) \in D} \psi(b, d) \int_b^d 1 dx
\end{align*}

since $\dmin \leq b$ and $d \leq \dmax$. Solving this integral, we obtain

\begin{equation}\label{eq:beta_0}
  \beta_0 = \frac{2}{A} \sum\limits_{(b, d) \in D} \psi(b, d) (d - b)
\end{equation}

Similarly to Equation~\eqref{eq:beta_0}, we can derive a closed-form formula for
$\beta_n$, $n \geq 1$.

\begin{align}
  \beta_n &= \frac{2}{A} \int_{\dmin}^{\dmax} \sum\limits_{(b, d) \in D} \psi(b, d) \chi_{[b,
  d)}(x) e^{2 \pi i n x / L} dx \nonumber \\
          &= \frac{2}{A} \sum\limits_{(b, d) \in D} \psi(b, d) \int_b^d e^{2 \pi
          i n x / A} dx \nonumber \\
  \beta_n &= \frac{2}{A}\sum\limits_{(b, d) \in D} \psi(b, d) \left[\frac{i A}{2
          \pi n}\left(e^{2 \pi i b n / A} - e^{2 \pi i d n / A}\right)\right]
          \label{eq:beta_n}
\end{align}

Therefore, from Equation~\eqref{eq:beta_0} and Equation~\eqref{eq:beta_n}, we
can calculate an arbitrary number of Fourier series coefficients for the
continuous persistence curve function.

To reconstruct the persistence curve from these Fourier series coefficients, we
implement the Equation~\eqref{eq:fapc_reconst} on a sampled grid $x$. We suggest
using $x = [\dmin, \dmax]$ with $1000$ samples.

\begin{equation} \label{eq:fapc_reconst}
  \begin{split}
    P(D)(x) &= \frac{\beta_0}{2} + \sum\limits_{n=1}^N [\Re\{\beta_n\} \cos(\frac{2 \pi n x}{A})  \\
            &+ \Im\{\beta_n\} \sin(\frac{2 \pi n x}{A})]
  \end{split}
\end{equation}

\section{Constants for HEPC and SP-FAPC Fitting}\label{apd:fit_constants}

For HEPC fitting, we scale each diagram by values in
Table~\ref{tab:scaling_params}. For SP-FAPC fitting, we use $\dmin$ and $\dmax$
values as determined by the domain specified in Table~\ref{tab:scaling_params}.
HEPC scaling values are chosen by average of 5 divided by the maximum
non-infinity value of each PD, calculated using a 10000 randomly selected
persistence diagrams. The SP-FAPC domain is chosen by caculating the 25th and
75th percentiles of the minimum and maximum of the PC domain respectively, using
the same randomly chosen PDs as the HEPC coefficient selection algorithm.

\begin{table}[htbp]
  \floatconts
  {tab:scaling_params}
  {\caption{Scaling values for HEPC fitting and set domain used for SP-FAPC
  fitting}}
  {
    \resizebox{\linewidth}{!}{
      \begin{tabular}{| c || c | c |}
        \hline
        \textbf{PD Type} & \textbf{HEPC Scaling Value} & \textbf{SP-FAPC Domain} \\
        \hline
        \hline
        $H_0$ Rips Airflow & 90442.544 & [0, 0.0002] \\
        \hline
        $H_1$ Rips Airflow & 55034.829 & [0, 0.0005] \\
        \hline
        $H_0$ Sublevel Airflow & 15909.436 & [-0.0015, 0.0015] \\
        \hline
        $H_0$ Sublevel IRR & 0.164 & [10, 50] \\
        \hline
      \end{tabular}
    }
  }
\end{table}

\section{XGBoost Parameters}\label{apd:xgboost_params}

We include a table of XGBoost parameters used for all models trained in this
paper in Table~\ref{tab:xgboost_params}.

\begin{table}[htbp]
  \centering
  \floatconts{tab:xgboost_params}
  {\caption{XGBoost Parameters}}
  {
    \begin{tabular}{| c | c |}
      \hline
      \textbf{Parameter} & \textbf{Value} \\
      \hline
      \hline
      learning\_rate & 0.07 \\
      n\_estimators & 100 \\
      max\_depth & 4 \\
      min\_child\_weight & 1 \\
      gamma & 0 \\
      subsample & 0.3 \\
      colsample\_bytree & 0.8 \\
      objective & multi:softprob \\
      eval\_metric & mlogloss \\
      seed & 999 \\
      \hline
    \end{tabular}
  }
\end{table}

\section{Confusion Matrix}\label{apd:confusion_matrix}

We present a confusion matrix for our best-performing model which uses Baseline,
AP-FAPC, and HEPC features in Table~\ref{tab:confusion_matrix}

\begin{table}[htpb]
  \centering
  \floatconts{tab:confusion_matrix}
  {\caption{Confusion Matrix for XGBoost Model Using Baseline + AP-FAPC + HEPC Features}}
  {
    \resizebox{\linewidth}{!}{
      \begin{tabular}{| c | c | c | c || c |}
        \hline
              & Awake   & NREM    & REM     & Class Accuracies\\
        \hline
        Awake & 9171.0  & 1580.6  & 2016.0  & 0.718 \\
        \hline
        NREM  & 21687.4 & 83967.8 & 20542.2 & 0.665 \\
        \hline
        REM   & 4797    & 2975    & 14101.2 & 0.645 \\
        \hline
      \end{tabular}
    }
  }
\end{table}

\section{Demographics Results and Sample Counts}\label{apd:demographics}

We include a table of subject counts for each demographic group used for age/sex
matching in the 5-fold validation and the associated average test balanced accuracy
for the XGBoost model on that demographic across averaged across all 5 folds in
Table~\ref{tab:demographics}. We include a table of sample counts for each class and the
associated percentage of the total dataset that class represents in
Table~\ref{tab:sample_counts}.

\begin{table}[htbp]
  \centering
  \floatconts{tab:demographics}
  {\caption{Demographic Counts and Balanced Accuracy (BA) using an XGBoost model
  trained on basline, HEPC, and AP-FAPC features combined}}
  {
    \begin{tabular}{| c | c | c | c |}
      \hline
      \textbf{Demographic} & \textbf{Subject Count} & \textbf{Train BA} &
      \textbf{Test BA}\\
      \hline
      \hline
      Age 2 Male & 63 & 0.674 & 0.650 \\
      \hline
      Age 2 Female & 32 & 0.669 & 0.644\\
      \hline
      Age 3 Male & 56 & 0.697 & 0.682 \\
      \hline
      Age 3 Female & 41 & 0.683 & 0.670 \\
      \hline
      Age 4 Male & 51 & 0.695 & 0.672 \\
      \hline
      Age 4 Female & 34 & 0.714 & 0.697 \\
      \hline
      Age 5 Male & 48 & 0.691 & 0.672 \\
      \hline
      Age 5 Female & 36 & 0.736 & 0.723 \\
      \hline
      Age 6 Male & 56 & 0.676 & 0.670 \\
      \hline
      Age 6 Female & 43 & 0.753 & 0.744 \\
      \hline
      Age 7 Male & 58 & 0.715 & 0.703 \\
      \hline
      Age 7 Female & 43 & 0.735 & 0.715 \\
      \hline
      Age 8 Male & 58 & 0.686 & 0.646 \\
      \hline
      Age 8 Female & 33 & 0.729 & 0.721 \\
      \hline
      Age 9 Male & 45 & 0.691 & 0.669 \\
      \hline
      Age 9 Female & 27 & 0.741 & 0.728 \\
      \hline
      Age 10 Male & 29 & 0.714 & 0.704 \\
      \hline
      Age 10 Female & 22 & 0.710 & 0.695 \\
      \hline
      Age 11 Male & 33 & 0.710 & 0.702 \\
      \hline
      Age 11 Female & 32 & 0.668 & 0.653 \\
      \hline
      Age 12 Male & 31 & 0.713 & 0.681 \\
      \hline
      Age 12 Female & 26 & 0.704 & 0.677 \\
      \hline
      Age 13 Male & 40 & 0.662 & 0.660 \\
      \hline
      Age 13 Female & 26 & 0.689 & 0.654 \\
      \hline
      Age 14 Male & 19 & 0.693 & 0.649 \\
      \hline
      Age 14 Female & 26 & 0.704 & 0.673 \\
      \hline
      Age 15 Male & 30 & 0.718 & 0.687 \\
      \hline
      Age 15 Female & 29 & 0.705 & 0.684 \\
      \hline
      Age 16 Male & 16 & 0.648 & 0.633 \\
      \hline
      Age 16 Female & 29 & 0.681 & 0.658 \\
      \hline
      Age 17 Male & 19 & 0.667 & 0.623 \\
      \hline
      Age 17 Female & 24 & 0.676 & 0.653 \\
      \hline
    \end{tabular}
  }

  \floatconts{tab:sample_counts}
  {\caption{Sample Counts for Dataset}}
  {
    \begin{tabular}{| c | c | c |}
      \hline
      \textbf{Label} & \textbf{Count} & \textbf{\% Of Dataset}\\
      \hline
      \hline
      Awake & 58272 & 7.8 \% \\
      \hline
      NREM Sleep  & 583261 &   78.6 \% \\
      \hline
      REM Sleep & 100895 & 13.6 \% \\
      \hline
    \end{tabular}
  }
\end{table}

\end{document}